\documentclass[11pt,a4paper]{article}
\usepackage[hyperref]{emnlp-ijcnlp-2019}
\usepackage{times}
\usepackage{latexsym}

\usepackage{url}

\usepackage{times}
\usepackage{latexsym}
\usepackage{booktabs}
\usepackage{comment}
\usepackage{footnote}
\usepackage{microtype}
\usepackage{amsmath}
\usepackage{graphicx}
\usepackage{cleveref}
\usepackage{tikz}
\usepackage{etoolbox}
\usepackage{enumitem}
\usepackage{xcolor}

\aclfinalcopy 

\title{Quick and (not so) Dirty: Unsupervised Selection of Justification Sentences for Multi-hop Question Answering}

\author{Vikas Yadav, Steven Bethard, Mihai Surdeanu \\
  University of Arizona, Tucson, AZ, USA \\
  {\tt \{vikasy, bethard, msurdeanu\}@email.arizona.edu}}
  
\date{}

\begin{document}
\maketitle
\begin{abstract}


We propose an unsupervised strategy for the selection of justification sentences for multi-hop question answering (QA) that 
(a) maximizes the relevance of the selected sentences,
(b) minimizes the overlap between the selected facts, and 
(c) maximizes the coverage of both question and answer. 
This unsupervised sentence selection method can be coupled with any supervised QA approach.
We show that the sentences selected by our method improve the performance of a state-of-the-art supervised QA model on two multi-hop QA datasets: AI2's Reasoning Challenge (ARC) and Multi-Sentence Reading Comprehension (MultiRC). 
We obtain new state-of-the-art performance on both datasets among approaches that do not use external resources for training the QA system: 56.82\% F1 on ARC (41.24\% on Challenge and 64.49\% on Easy) and 26.1\% EM0 on MultiRC.
Our justification sentences have higher quality than the justifications selected by a strong information retrieval baseline, e.g., by 5.4\% F1 in MultiRC.
We also show that our unsupervised selection of justification sentences is more stable across domains than a state-of-the-art supervised sentence selection method. 

  
  
\end{abstract}

\section{Introduction}


Interpretable machine learning (ML) models, where the end user can understand how a decision was reached, are a critical requirement for the wide adoption of ML solutions in many fields such as healthcare, finance, and law~\cite{samek2017explainable, alvarez2017causal, arras2017relevant, gilpin2018explaining, biran2017explanation} 

For complex natural language processing (NLP) such as question answering (QA),  human readable explanations of the inference process have been proposed as a way to interpret QA models~\cite{zhou2018interpretable}.

\begin{figure}[t!]
\small
To which organ system do the esophagus, liver, pancreas, small intestine, and colon belong?

(A) reproductive system (B) excretory system \\
(C) {\bf digestive system} (D) endocrine system

\hrulefill

ROCC-selected justification sentences:
\begin{enumerate}[nosep]
\item vertebrate {\bf digestive system} has oral cavity, teeth and pharynx, \emph{esophagus} and stomach, \emph{small intestine, pancreas, liver} and the large intestine
\item {\bf digestive system} consists liver, stomach, large intestine, \emph{small intestine, colon}, rectum and anus
\end{enumerate}

\hrulefill

BM25-selected justification sentences:
\begin{enumerate}[nosep]
\item their {\bf digestive system} consists of a stomach, \emph{liver}, \emph{pancreas}, \emph{small intestine}, and a large intestine
\item the \emph{liver pancreas} and gallbladder are the solid organ of the {\bf digestive system}
\end{enumerate}
\caption{\label{fig:coverage_example}\small A multiple-choice question from the ARC dataset with the correct answer in bold, followed by justification sentences selected by our approach (ROCC) vs. sentences selected by a strong IR baseline (BM25). 
ROCC justification sentences fully cover the five key terms in the question (shown in italic), whereas BM25 misses two: {\em esophagus} and {\em colon}. Further, the second BM25 sentence is largely redundant with the first, not covering other query terms.
}
\end{figure}

Recently, multiple datasets have been proposed for {\em multi-hop} QA, in which questions can only be answered when considering information from multiple sentences and/or documents~\cite{clark2018think, khashabi2018looking, yang2018hotpotqa, welbl2018constructing, mihaylov2018can, bauer2018commonsense, dunn2017searchqa, dhingra2017quasar, lai2017race, rajpurkar2018know, sun2019dream}. 
The task of selecting justification sentences is complex for multi-hop QA, because of the additional knowledge aggregation requirement (examples of such questions and answers are shown in Figures~\ref{fig:coverage_example} and~\ref{fig:ROCC_search_process}).
Although various neural QA methods have achieved high performance on some of these datasets~\cite{sun2018improving, trivedi2019repurposing, Tymoshenko2017RankingKF, seo2016bidirectional, wang2016compare, de2018question,back2018memoreader}, 
we argue that more effort must be dedicated to explaining their inference process.

In this work we propose an {\em unsupervised} algorithm for the selection of {\em multi-hop} justifications from {\em unstructured} knowledge bases (KB). 
Unlike other supervised selection methods~\cite{dehghani2019learning, bao2016constraint, lin2018denoising, wang2018multi, wang2018evidence,tran2018multihop, trivedi2019repurposing}, our approach does not require any training data for justification selection. Unlike approaches that rely on structured KBs, which are expensive to create, \cite{khashabi2016question, khot2017answering, Zhang2018KG2LT, khashabi2018question, cui2017kbqa, bao2016constraint}, our method operates over KBs of only unstructured texts. We demonstrate that our approach has a bigger impact on downstream QA approaches that use these justification sentences as additional signal than a strong baseline that relies on information retrieval (IR). 
In particular, the contributions of this work are:
\begin{enumerate}[label={\bf(\arabic*)}, topsep=0.2em, itemsep=0.1em, wide, labelwidth=!, labelindent=0pt]
\item We propose an unsupervised, non-parametric strategy for the selection of justification sentences for multi-hop question answering (QA) that 
(a) maximizes the {\bf {\em R}}elevance of the selected sentences; 
(b) minimizes the lexical {\bf {\em O}}verlap between the selected facts; and 
(c) maximizes the lexical {\bf {\em C}}overage of both question and answer. We call our approach ROCC. 
ROCC operates by first creating $n \choose k$ justification sets from the top $n$ sentences selected by the BM25 information retrieval model~\cite{BM25}, where $k$ ranges from 2 to $n$, and then ranking them all by a formula that combines the three criteria above. The set with the top score becomes the set of justifications output by ROCC for a given question and candidate answer.
As shown in Figure~\ref{fig:coverage_example}, 
 the justification sentences selected by ROCC perform more meaningful knowledge aggregation than a strong IR baseline (BM25), which does not account for overlap (or complementarity) and coverage. 

\item ROCC can be coupled with any supervised QA approach that can use the selected justification sentences as additional signal. To demonstrate its effectiveness, we combine ROCC with a state-of-the-art QA method that relies on BERT~\cite{devlin2018bert} to classify correct answers, using the text of the question, the answer, and (now) the justification sentences as input. 
On the Multi-Sentence Reading Comprehension (MultiRC) dataset~\cite{khashabi2018looking}, we achieved a gain of 8.3\% EM0 with ROCC justifications when compared to the case where the complete comprehension passage was provided to the BERT classifier. 
On AI2's Reasoning Challenge (ARC) dataset~\cite{clark2018think}, the QA approach enhanced with ROCC justifications outperforms the QA method without justifications by 9.15\% accuracy, and the approach that uses top sentences provided by BM25 by 2.88\%.  
Further, we show that the justification sentences selected by ROCC are considerably more correct on their own than justifications selected by BM25 (e.g., the justification score in MultiRC was increased by 11.58\% when compared to the best performing BM25 justifications), %
which indicates that the interpretability of the overall QA system was also increased. 

\item Lastly, our analysis indicates that ROCC is more stable across the different domains in the MultiRC dataset than a supervised strategy for the selection of justification sentences that relies on a dedicated BERT-based classifier, 
with a difference of over 10\% F1 score in some configurations. 
\end{enumerate}

The ROCC system and the codes for generating all the analysis are provided here - \url{https://github.com/vikas95/AutoROCC}.

\section{Related Work}

The body of QA work that addresses the selection of justification sentences can be classified into roughly four categories: (a) supervised approaches that require training data to learn how to select justification sentences (i.e., questions and answers coupled with correct justifications); (b) methods that treat justifications as latent variables and learn jointly how to answer questions and how to select justifications from questions and answers alone; (c) approaches that rely on information retrieval to select justification sentences; and, lastly, (d) methods that do not use justification sentences at all. 

In the first category, previous works (e.g.,~\cite{trivedi2019repurposing}) have used entailment resources including labeled trained datasets such as  SNLI \cite{bowman2015large} and MultiNLI \cite{williams2017broad} to train components for selecting justification sentences for QA. Other works have explicitly focused on training sentence selection components for QA models \cite{min2018efficient, lin2018denoising, wang2019evidence}. In datasets where gold justification sentences are not provided, researchers have trained such components by retrieving justifications from structured KBs \cite{cui2017kbqa, bao2016constraint, zhang2016joint, hao2017end} such as ConceptNet \cite{speer2017conceptnet}, or from IR systems coupled with denoising components \cite{wang2019evidence}.
While these works offer exciting directions, they all rely on training data for justifications, which is expensive to generate and may not be available in real-world use cases.

The second group of methods tend to rely on reinforcement learning \cite{choi2017coarse, lai2018review, geva2018learning} or PageRank \cite{surdeanu2008learning} to learn how to select justification sentences without explicit training data. Other works have used end-to-end (mostly RNNs with attention mechanisms) QA architectures for learning to pay more attention on better justification sentences~\cite{min2018efficient, seo2016bidirectional, yu2014deep, gravina2018cross}.
While these approaches do not require annotated justifications, they need large amounts of question/answer pairs during training so they can discover the latent justifications. In contrast to these two directions, our approach requires no training data at all for the justification selection process. 

The third category of methods utilize IR techniques to retrieve justifications from both unstructured \cite{yadav_AHE} and structured \cite{khashabi2016question} KBs. 
Our approach is closer in spirit to this direction, but it is adjusted to account for more intentional knowledge aggregation. As we show in Section~\ref{sec:results}, this is important for both the quality of the justification sentences and the performance of the downstream QA system.

The last group of QA approaches learn how to classify answers without any justification sentences~\cite{mihaylov2018can, sun2018improving, devlin2018bert}. While this has been shown to obtain good performance for answer classification, we do not focus on it in this work because these methods cannot easily explain their inference. 

Note that some of the works discussed here transfer knowledge from external datasets into the QA task they address~\cite{chung2017supervised, sun2018improving, pan2019improving, min2017question, qiu2018transfer, chen2017reading}. In this work, we focus solely on the resources provided in the task itself because such compatible external resources may not be available in real-world applications of QA.

\section{Approach}

\begin{figure}[t!]
\includegraphics[width=\columnwidth]{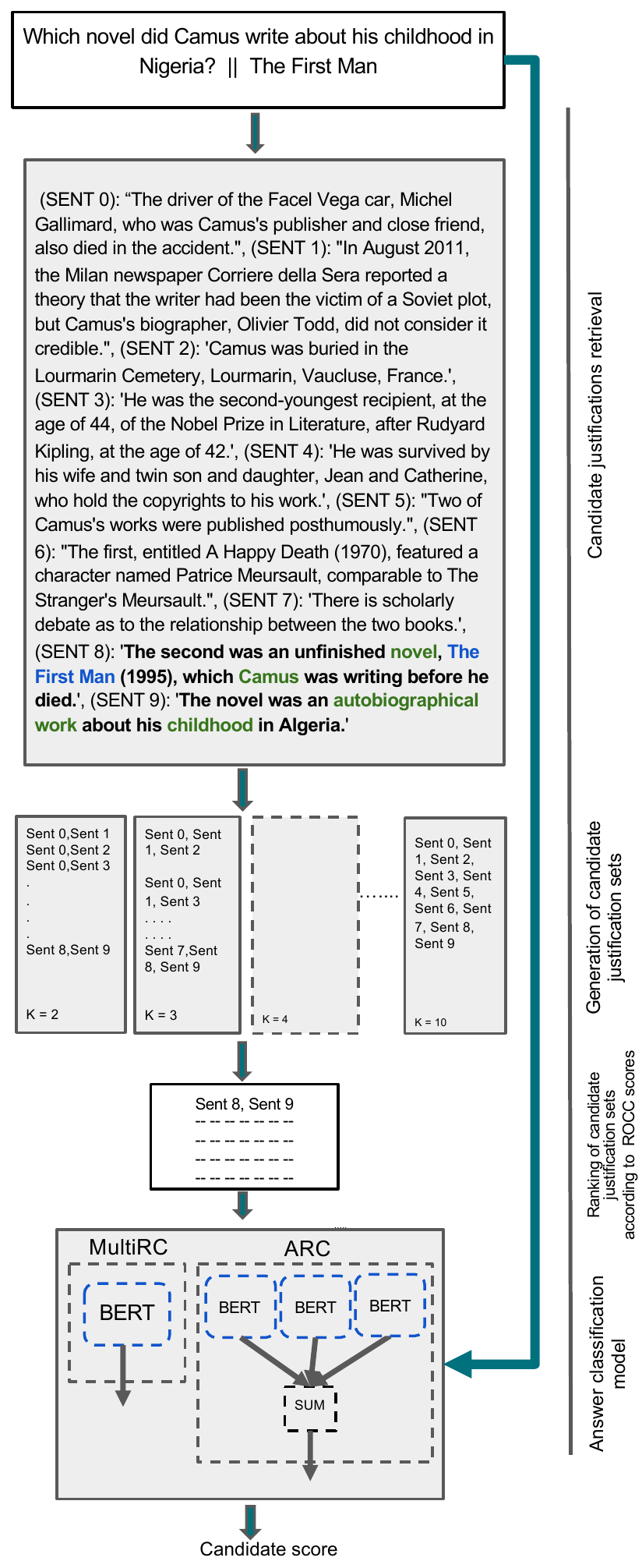}  
\vspace{-\baselineskip}
\caption{\label{fig:ROCC_search_process}
An example of the ROCC process for a question from the MultiRC dataset. Here, ROCC correctly extracts the two justification sentences necessary to explain the correct answer.
 }
\end{figure}

ROCC, coupled with a QA system, operates in the following steps (illustrated in Figure~\ref{fig:ROCC_search_process}):

\begin{enumerate}[label={\bf(\arabic*)}, topsep=0.2em, itemsep=0.1em, wide, labelwidth=!, labelindent=0pt]
\item \textbf{Retrieval of candidate justification sentences}: For datasets that rely on huge supporting KBs (e.g., ARC), we retrieve the top $n$ sentences\footnote{In this work we used $n = 20$ as in \citet{yadav_AHE}} from this KB using an IR query that concatenates the question and the candidate answer, similar to \citet{clark2018think, yadav_AHE}. We implemented this using the BM25 IR model with the default parameters in  Lucene\footnote{\url{https://lucene.apache.org}}.
For reading comprehension datasets where the question is associated with a text passage (e.g., MultiRC), all the sentences in this passage become candidates. 

\item \textbf{Generation of candidate justification sets}: Since its focus is on knowledge aggregation, ROCC ranks {\em sets} of justification sentences (see below) rather than individual sentences. In this step we create candidate justification sets by generating $n \choose k$ groups of sentences from the previous $n$ sentences, using multiple values of $k$. 

\item \textbf{Ranking of candidate justification sets}:  For every candidate justification set, we calculate its ROCC score (see Section~\ref{ROCC_ranking}), which estimates the likelihood that this group of justifications explains the given answer. We then rank the justification sets in descending order of ROCC score, and choose the top set as the group of justifications that is the output of ROCC for the given question and answer. In MultiRC, we rearrange the justification sentences according to their original indexes in the given passage to bring coherence in the selected sequence of sentences.

\item \textbf{Answer classification}:  
ROCC can be coupled with any supervised QA component for answer classification. 
In this work, we feed in the question, answer, and justification texts into a state-of-the-art classifier that relies on BERT (see Section~\ref{sec:bert}). 
Because the justification sentences in the reading comprehension use case (e.g., MultiRC) come from the same passage and their sequence is likely to be coherent, we concatenate them into a single passage, and use a single BERT instance for classification. 
This approach is shown on the left side of the answer classification component in Figure~\ref{fig:ROCC_search_process}.
On the other hand, the justification sentences retrieved from an external KB (e.g., ARC) may not form a coherent passage when aggregated. For this reason, in the ARC use case, we classify each justification sentence separately (together with the question and candidate answer), and then average all these scores to produce a single score for the candidate answer (right-hand side of the figure).
\end{enumerate}

%

\vspace{-3mm}

\subsection{Ranking of Candidate Justification Sets}
\label{ROCC_ranking}
%
%

Each set of justifications is ranked based on its ROCC score, which:
(a) maximizes the {\em R}elevance of the selected sentences; 
(b) minimizes the lexical {\em O}verlap between the selected facts; and 
(c) maximizes the lexical {\em C}overage of both question and answer ($C_{ques}, C_{ans}$).
The overall score for a given justification set $P_i$ is calculated as:
\begin{equation}
    S(P_i) = \frac{R}{\epsilon+O(P_i)} \cdot (\epsilon+C(A)) \cdot (\epsilon+C(Q))
    \label{ROCC_formula}
\end{equation} 
To avoid zeros, we add a small constant ($\epsilon=1$ here) to each component that can have a value of 0.\footnote{Our R score relies on BM25, which is larger than 0 on the top $n$ sentences.}
We detail the components of this formula below. 

\begin{description}[topsep=0.2em, itemsep=0.1em, wide, labelwidth=!, labelindent=0pt]

\item[Relevance (R)] We use the Lucene implementation\footnote{\url{https://lucene.apache.org/core/7_0_1/core/org/apache/lucene/search/similarities/BM25Similarity.html}} of the BM25 IR model~\cite{BM25} to estimate the relevance of each justification sentence to a given question and candidate answer.
In particular, we form a query that concatenates the question and candidate answer, and use as underlying document collection (necessary to compute document statistics such as inverse document frequencies (IDF)) either: sentences in the entire KB (for ARC), or all sentences in the corresponding passage in the case of reading comprehension (MultiRC). 
The arithmetic mean of BM25 scores over all sentences in a given justification set gives the value of $R$ for the entire set.

%

\item[Overlap (O)] To ensure diversity and complementarity between justification sentences, we compute the overlap between all sentence pairs in a given group. Thus, minimizing this score reduces redundancy and encourages the aggregated sentences to address {\em different} parts of the question and answer:
\begin{equation}
O(S) = \dfrac{\displaystyle\sum_{s_i \in S} \sum_{s_j \in S - s_i} \dfrac{|t(s_i) \cap t(s_j)|}{\max (|t(s_i)|, |t(s_j)|)}} { {|S| \choose 2}}
\end{equation}
where $S$ is the given set of justification sentences;
$s_i$ is the $i^{th}$ sentence in $S$; and $t(s_i)$ denotes the set of unique terms in sentence $s_i$.
Note that we divide by $|S| \choose 2$ to normalize across different sizes of justification sets. 


\item[Coverage (C)]
Complementing the overlap score, this component measures the lexical coverage of the question and the answer texts by the given set of justifications $S$. This coverage is weighted by the IDF of question and answer terms. Thus, maximizing this value encourages the justifications to address more of the meaningful content mentioned in the question ($X=Q$) and the answer ($X=A$):
\begin{align}
    C_t(X) &= \displaystyle\bigcup_{s_i \in S} t(X) \cap t(s_i)\\
    C(X) &= \dfrac{\sum_{t=1}^{|C_t(X)|} IDF[C_t(X)[t]]}{|t(X)|}
    \label{Coverage_IDF}
\end{align}
where $t(X)$ denotes the unique terms in $X$, and $C_t(X)$ represents the set of all unique terms in $X$ that are present in any of the sentences of the given justification set. $C(X)$ gives the IDF weighted average of $C_t(X)$ terms.

\end{description}

\subsection{Answer Classification}
\label{sec:bert}

As indicated earlier, we propose two flavors for the answer classification component: if the sentences in a justification group come from the same passage and, thus, are likely to be coherent, they are concatenated into a single text before classification, and handled by a single answer classifier. If the sentences come from different texts, they are handled by separate instances of the answer classifier. In the latter case, all scores are averaged to produce a single score for a candidate answer. 
In all situations we used BERT \cite{devlin2018bert} for answer classification. In particular, we employed BERT as a binary classifier operating over two texts.  The first text consists of the concatenated question and answer, and the second text consists of the justification text. The classifier operates over the hidden states of the two texts, i.e., the state corresponding to the {\tt [CLS]} token \cite{devlin2018bert}.\footnote{We used the following hyper parameters with BERT Large: learning rate of 1e-5, maximum sequence length of 128, batch size = 16, number of epochs = 6.} 


We observed empirically that pre-training the BERT classifier on {\em all} $n$ sentences retrieved by BM25, and then fine tuning on the ROCC justifications improves performance on all datasets we experimented with. This resembles the transfer learning discussed by~\citet{howard2018universal}, where the source domain would be the BM25 sentences, and the target domain the ROCC justifications. However, one important distinction is that, in our case, all this knowledge comes solely from the resources provided within each dataset, and is retrieved using unsupervised method (BM25). We conjecture that this helped mainly because the pre-training step exposed BERT to more data which, even if imperfect, is topically related to the corresponding question and answer.

\begin{table}[]
    \centering
    \footnotesize{}
    \begin{tabular}{l|l}
        Question + answer text & Justification set  \\
        \hline
       Animal cells obtain energy  & 1) obtain water and nutrient by \\
     by $||$ absorbing nutrients  & absorbing them directly into \\
     & plant cell \\
     
     & 2) the animal obtain nourish- \\
     & ment by absorbing nutrient  \\
     & released by symbiotic bacteria 
       
    \end{tabular}
    \caption{\small Example of a justification set in ARC which was scored by annotator with a precision of $\frac{1}{2}$ because the first justification sentence is not relevant, and a coverage of $\frac{1}{2}$ because the link between {\em nourishment} and {\em energy} is not covered.}
    \label{tab:ARC_justification_analysis}
\end{table}

\begin{table*}[t]
\centering
\footnotesize
\setlength{\tabcolsep}{0.25em}
\begin{tabular}{@{} lllllllccc @{}}
\hline
\# & External  & Supervised & Method &  F1$_m$ & F1$_a$ & EM0 & \multicolumn{3} {c}{Justification}   \\ 
  & resource? & selection of & & & & & P & R & F1 \\
  & & justifications? \\
 \hline
 & & & \bf{DEVELOPMENT DATASET} \\
 \hline 
 & & & Baselines \\
 \hline 
 0 & No & No & Predict 1 \cite{khashabi2018looking} & 61.0 & 59.9 & 0.8 &  -- \\
 1 & No & No & IR(paragraphs) \cite{khashabi2018looking} & 64.3 & 60.0 & 1.4 &  -- \\
 2 & No & No & SurfaceLR \cite{khashabi2018looking} & 66.5 & 63.2 & 11.8 &  -- \\
 3 & No & No & Entailment baseline \cite{trivedi2019repurposing} & 51.3 & 50.4 & -- & -- \\
\hline
& & & Previous work \\
 \hline 
4 & Yes & Yes & EER$_{DPL}$ + FT \citet{wang2019evidence} & 70.5 & 67.8 & 13.3 &  -- \\
5 & Yes & Yes & Multee (GloVe) \cite{trivedi2019repurposing} & 71.3 & 68.3 & 17.9 & -- \\
6 & No & Yes & Multee (ELMo) \cite{trivedi2019repurposing} & \bf{\textit{70.3}} & 67.3 & \bf{\textit{22.8}} &  -- \\
7 & Yes & Yes & Multee (ELMo) \cite{trivedi2019repurposing} & 73.0 & 69.6 & 22.8 &  -- \\
8 & No & Yes & RS \cite{sun2018improving} & 69.7 & \bf{\textit{67.9}} & 16.9 & --\\
9 & Yes & Yes & RS \cite{sun2018improving} & \bf{73.1}$^\star$ & \bf{70.5}$^\star$  & 21.8 &-- \\
\hline
& & & BERT + IR baselines \\
	 \hline 
	10 & No & No & BERT + entire passage & 65.7 & 62.7 & 17.0 &17.4 & 100.0 & 29.6 \\
	11 & No & No & BERT + BM25 ($k = 1$ sentence) & 66.2 & 62.8 & 17.9 & {\bf61.0} & 27.1 & 37.5 \\
	12 & No & No & BERT + BM25 ($k = 2$ sentences) & 68.1 & 64.8 & 21.0 &  51.6 & 45.6 & 48.4 \\
	13 & No & No & BERT + BM25 ($k = 3$ sentences) & 69.1 & 65.7  & 21.6 &  42.6 & 56.1 & 48.4 \\
	14 & No & No & BERT + BM25 ($k = 4$ sentences) & 70.05 & 66.7 & 22.3 &  36.9 & 64.6 & 47.0 \\
	15 & No & No & BERT + BM25 ($k = 5$ sentences) & 71.2 & 67.7 & 23.4 &  32.7 & 71.1 & 44.8 \\
\hline
& & & BERT + parametric ROCC \\
	 \hline 
	16 & No & No & BERT + ROCC ($k = 2$ sentences) & 69.8 & 66.8 & 22.7 &  54.7 & 48.5 & 51.4 \\
	17 & No & No & BERT + ROCC ($k = 3$ sentences) & 72.7 & {\bf69.7} & 25.2 &  48.0 & 63.5 & 54.7 \\
	18 & No & No & BERT + ROCC ($k = 4$ sentences) & 72.2 & 69.0 & 25.0 & 40.6 & 71.0 & 51.6\\
	19 & No & No & BERT + ROCC ($k = 5$ sentences) & 71.6 & 68.7 & 22.7 & 35.0 & {\bf76.5} & 48.1\\
\hline
& & & BERT + non-parametric ROCC \\
	 \hline
	20 & No & No & BERT + AutoROCC ($k \in \{2, 3, 4\}$) & 72.0 & 69.0 & 21.9  & 48.9 & 66.5 & 56.3 \\
	21 & No & No & BERT + AutoROCC ($k \in \{2, 3, 4, 5\}$) & 72.0 & 68.8 & 23.5  & 48.3 & 67.7 & 56.4\\
	22 & No & No & BERT + AutoROCC ($k \in \{2, 3, 4, 5, 6\}$) & 72.1 & 69.2 & {\bf25.3} & 48.2 & 68.2 & {\bf56.4} \\
	23 & No & No & BERT + BM25 ($k$ from best AutoROCC) & 71.1 & 67.4 & 23.1 & 43.8 & 61.2 & 51.0 \\
	24 & No & No & BERT + AutoROCC ($k \in \{2, 3, 4, 5, 6\}$, pre-trained) & {\bf72.9} & 69.6 & 24.7 & 48.2 & 68.2 & {\bf56.4}\\
\hline
& & & Ceiling systems with gold justifications \\
	 \hline
25 & Yes & Yes & EER$_{gt}$ + FT \cite{wang2019evidence} & 72.3 & 70.1 & 19.2 & - -\\
26 & No & Yes & BERT + Gold knowledge & 79.1 & 75.4 & 37.6 &  100.0 & 100.0 & 100.0\\
27 & - & - & Human & 86.4 & 83.8 & 56.6 &  -- \\
\hline
& & & \bf{TEST DATASET} \\
\hline
28 & No & No & SurfaceLR \cite{khashabi2018looking} & 66.9 & 63.5 & 12.8\\
29 & Yes & Yes & Multee (ELMo) \cite{trivedi2019repurposing} & 73.8 & 70.4 & 24.5 & -- \\
30 & No & No & BERT + AutoROCC ($k \in \{2, 3, 4, 5, 6\}$, pre-trained) & {\bf73.8} & {\bf70.6} & {\bf26.1}  \\
\end{tabular}
\vspace{-0.5\baselineskip}
\caption{\small Performance on the MultiRC dataset, under various configurations. 
	$k$ indicates the size(s) of the sets of justification sentences. 
	In parametric ROCC, $k$ is a hyper parameter; in AutoROCC, $k$ is selected automatically. 
	The pre-trained ROCC configurations pre-train BERT on the entire passage corresponding to the question, before fine tuning it on the ROCC sentences. 
	Bold values with $^\star$ indicate state-of-the-art results that used external labeled resources or other supervised methods for the selection of justification sentences. 
	Italicized bold values show state-of-the-art results from experiments that do not use any external labeled resources.
} 
\label{tab:MultiRC}
\end{table*}


\begin{table*}[t]
\centering
\footnotesize
\setlength{\tabcolsep}{0.25em}
\begin{tabular}{@{} lllllllll @{}}
\hline
\# & External  & Supervised & Method &  Challenge & Easy & All  & Justification  \\ 
	 & resources & selection of & & & & & P, Coverage  \\
	 & used? & justifications? & \\
\hline
 & & & Baselines \\
\hline

0	& No & No &	AI2 IR Solver \citep{clark2018think}	& 59.99 & 23.98 & & $>0$ \\
1	& No & No &	Sanity Check  \cite{yadav2018sanity} & 	58.36 & 26.56 & & $>0$ \\
2 & Yes & No & Tuple-Inf \cite{clark2018think}& 60.71 & 23.83 & & $>0$ \\


3 & Yes & No & DGEM \cite{clark2018think}& 58.97 & 27.11 & & $>0$\\ 


\hline
& & & Previous work \\
\hline


4 & Yes & -- & Bi-LSTM max-out  \cite{mihaylov2018can} & 33.87 & 34.26  &  & $=0$\\ 




8 & No & No & AHE \cite{yadav_AHE} & 33.28 & \bf{\textit{63.22}} & 53.31  \\

9 & No & -- & Reading Strategies \cite{sun2018improving} & \bf{\textit{35.40}} & 63.10 & \bf{\textit{53.94}} & $=0$ \\

10 & Yes & -- & Reading Strategies \cite{sun2018improving} & \bf{42.30}$^\star$ & \bf{68.90}$^\star$ & \bf{60.19}$^\star$ & $=0$ \\

\hline
  & & & BERT + IR baselines & & &  \\
\hline
11 & No & -- & BERT & 35.11 & 52.75 & 46.94  \\
	12 & No & No & BERT + BM25 ($k = 1$  sentence) & 33.87 & 56.23 & 48.85  \\
	13 & No & No & BERT + BM25 ($k = 2$  sentences) & 38.65 & 60.50 & 53.29  \\
	14 & No & No & BERT + BM25 ($k = 3$  sentences) & 41.04 & 63.19 & 55.89 \\
	15 & No & No & BERT + BM25 ($k = 4$  sentences) & 37.9 & 63.49 & 53.90  \\
	16 & No & No & BERT + BM25 ($k = 5$  sentences) & 38.01 & 61.28 & 53.60 \\
\hline
& & & BERT + parametric ROCC & & &  \\
\hline
17 & No & No & BERT + ROCC ($k = 2$  sentences) & 36.65 & 60.59 &  52.69 \\
	18 & No & No & BERT + ROCC ($k = 3$  sentences) & 39.29 & 62.97 & 55.16  \\
	19 & No & No & BERT + ROCC ($k = 4$  sentences) & 40.39 & 61.13 & 54.29  \\
	20 & No & No & BERT + ROCC ($k = 5$  sentences) & 40.62 & 59.96 & 53.58  \\
\hline
& & & BERT + non-parametric ROCC \\
	 \hline
	21 & No & No & BERT + AutoROCC ($k \in \{2, 3, ... 20  \}$) & 40.73 & 63.64 & 56.09 & 48.04, 62.50 \\
	22 & No & No & BERT + BM25 ($k$ from best AutoROCC) & 39.24 & 61.01 & 53.83 & 42.55, 55.88 \\
	23 & No & No & BERT + AutoROCC ($k \in \{2, 3, ... 20 \}$, pre-trained) & {\bf41.24} & {\bf64.49} & {\bf56.82} &  \bf{48.04, 62.50} \\
\end{tabular}
\vspace{-0.5\baselineskip}
\caption{Performance on the ARC dataset, under various configurations. Notations are the same as in Table~\ref{tab:MultiRC}.} 
\label{tab:ARC}
\end{table*}


\section{Empirical Evaluation}
\label{sec:results}

We evaluated ROCC coupled with the proposed QA approach on two QA datasets.
We use the standard train/development/test partitions for each dataset, as well as the standard evaluation measures: accuracy for ARC~\cite{clark2018think}, and F1${_m}$ (macro-F1 score), F1${_a}$ (micro-F1 score), and EM0 (exact match) 
for MultiRC~\cite{khashabi2018looking}.  

\paragraph{Multi-sentence reading comprehension (MultiRC): } this is a reading comprehension dataset implemented as multiple-choice QA~\cite{khashabi2018looking}. Each question is accompanied by a supporting passage, which contains the correct answer. We use all sentences from such paragraphs as candidate justifications for the corresponding questions.

\paragraph{AI2's Reasoning Challenge (ARC):} this is a multiple-choice question dataset, containing 
	questions from science exams from grade 3 to grade 9 \cite{clark2018think}. The dataset is split in two partitions: Easy and Challenge, where the latter partition contains the more difficult questions that require reasoning. 
	Most of the questions have 4 answer choices, with $<$1\% of all the questions having either 3 or 5 answer choices.
	Importantly, ARC includes a supporting KB of 14.3M unstructured text passages. We use BM25 over this entire KB to retrieve candidate justification sentences for ROCC. 

\subsection{Justification Results}
To demonstrate that ROCC has the capacity to select better justification sentences, we also report the quality of the extracted justification sentences. For MultiRC, we report precision/recall/F1 justification scores, computed against the gold justification sentences provided by the dataset.\footnote{We use these gold justifications only for evaluation, {\em not} for training, since ROCC is an unsupervised algorithm.}
For ARC, where gold justifications are not provided, we used an external annotator  to annotate the justifications for a random stratified sample of 70 questions, with 10 questions selected from each grade (3 -- 9).
The annotator reported two scores: precision,  and coverage. Precision was defined as the fraction of justification sentences that are relevant for the inference necessary to connect the corresponding question and candidate answer. Coverage was defined as 1 if the justification set completely covers the inference process for the given question and answer, $1/2$ if the set of justifications partially addresses the inference, and 0 if the justification set is completely irrelevant. Table~\ref{tab:ARC_justification_analysis} illustrates these scores with an actual output from ARC. 


\subsection{Question answering results}

In addition to comparing ROCC with previously reported results, we include multiple baselines: (a)  the BERT answer classifier trained on the entire passage 
of the given question (MultiRC), to demonstrate that ROCC has the capacity to filter out irrelevant content from these paragraphs; (b) BERT trained without any justification sentences (ARC), to show that ROCC has the capacity to aggregate useful information from large unstructured KBs, and (c) BERT trained on sentences retrieved using BM25, to demonstrate that ROCC performs better than other unsupervised approaches. Note that the BM25 baseline has an additional hyper parameter: the number of sentences to be considered ($k$). 


Table~\ref{tab:MultiRC} reports comprehensive results on MultiRC, including both overall QA performance, measured using F1$_m$, F1{$_a$}, and EM0, as well as justification quality, measured using standard precision (P), recall (R), and F1. Note that the bulk of the results are reported on the development partition. The last row in the table reports results on the test partition, computed using the official submission portal which can be accessed only once per model (including its variants). 
To understand ROCC's behavior, the table includes both the parametric form of ROCC, where the size of the justification sets ($k$) is manually tuned 
as well as the non-parametric ROCC, where $k$ is automatically selected in the third step of the ROCC algorithm (see Figure~\ref{fig:ROCC_search_process}) by sorting across all sizes of justification sets together, instead of sorting within each value of $k$. 
Table~\ref{tab:ARC} lists equivalent results on ARC.
	
\begin{table*}
    \centering
    \begin{tabular}{c|c|c|c|c|c|c|c}
    train/test    & Science  & Fiction & News & Wiki & wikiMovie& Society,  & All \\
    &  textbook &  &  & articles & Summaries & Law and \\
    & & & & & & Justice & \\
   \hline
   AutoROCC & 54.57 & 53.88 & 54.32 & 60.49 & 57.10 & 61.06 & 56.44 \\
    \hline
   BERT+All passages & 55.15 & 55.46 & 68.77 & 65.14 & 57.39 & 58.79 & 60.90\\   
   BERT+Science textbook & 55.67 & 41.01 & 51.45 & 50.06 & 54.96 & 48.84 & 50.79 \\
   BERT+Fiction & 45.16 & 57.60 & 63.05 & 63.13 & 59.98 & 50.94 &  58.31 \\
   BERT+News & 44.11 & 50.77 & 68.82 & 65.45 & 57.01 & 58.30 & 59.30 \\
   \hline
      GPT-2 \cite{wang2019evidence} & - & - & - & - & - & - & 60.7 \\
    \end{tabular}
    \caption{ Domain robustness of the non-parametric ROCC vs. a supervised sentence selection model, evaluated on the gold justification sentences from MultiRC.
    Each column represents a section of the MultiRC development set.
    Each row after AutoROCC represents a justification sentence selection component trained only on the specified section of MultiRC (these sections are listed in descending order of the number of passages in the training data).
    } 
    \label{tab:domain_robustness}
    \vspace{-4mm}
\end{table*}

We draw several observations from these tables:

\begin{enumerate}[label={\bf(\arabic*)}, topsep=0.2em, itemsep=0.1em, wide, labelwidth=!, labelindent=0pt]
\item Despite its simplicity, ROCC combined with the BERT classifier obtains new state-of-the-art performance on both MultiRC and ARC for the class of approaches that do not use external resources to either train the justification sentence selection or the answer classifier. For example, ROCC outperforms the previous best result in MultiRC by 2.5 EM0 points on the development partition (row 24 vs. row 6), and 1.6 EM0 points on test (row 30 vs. row 29). In ARC, ROCC outperforms the previous best approach by 5.8\% accuracy on the Challenge partition, and 2.9\% overall (row 23 vs. row 9). 

\item On both datasets, the non-parametric form of ROCC (AutoROCC) slightly outperforms the parametric variant. Importantly, it always achieves higher justification scores compared to the parametric ROCC.
In MultiRC, AutoROCC outperforms our baseline of BERT + entire passage (row 10 vs 22)  by 8.3\% EM0, indicating that AutoROCC can filter out irrelevant content. In ARC, AutoROCC outperforms the baseline with no justification sentences by 9.1\% (row 21 vs row 11), demonstrating that ROCC aggregates useful knowledge.

\item The results of the parametric forms of ROCC (rows 16 -- 19 in Table~\ref{tab:MultiRC} and rows 17 -- 20 in Table~\ref{tab:ARC}) indicate that performance continues to increase until $k = 4$ in MultiRC and $k = 3$ in ARC. This indicates that: (a) knowledge aggregation is beneficial for these tasks; (b) ROCC can robustly handle non-trivial cases of aggregation with larger values of $k$; and (c) similar to other QA methods~\cite{chen2019understanding}, performance decreases for large values of $k$, suggesting that knowledge aggregation remains an open research challenge. 

\item The justification scores in both datasets are considerably higher than the equivalent configuration that uses BM25 instead of ROCC (i.e., row 24 vs. row 23 in Table~\ref{tab:MultiRC}, and row 23 vs. row 22 in Table~\ref{tab:ARC}). 
This confirms that the {\em joint} scoring of sets of justifications that ROCC performs is better than the individual ranking of justification sentences performed by standard IR models such as BM25. 
\end{enumerate}


\subsection{Domain Robustness Analysis}
\label{domain_robustness}


To understand ROCC's domain robustness, we compared it against a supervised BERT-based classifier for the selection of justification sentences, as well as against GPT-2~\cite{wang2019evidence}.
 For this experiment, we used MultiRC, where gold justifications are provided. We used this data to train a classifier for the selection of justification sentences on various domain-specific sections of MultiRC. 
The results of this experiment are shown in Table~\ref{tab:domain_robustness}. Unsurprisingly, training and testing in the same domain (e.g., Fiction) leads to the best performance on sentence selection. However, ROCC is more stable across domains than the supervised sentence selection component, with a difference of over 10 F1 points in some configurations. This suggests that ROCC is a better solution for real-world use cases where the distribution of the test data may be very different from the training data.  

Compared to BERT, the unsupervised AutoROCC achieves almost the same or better performance in the majority of the domains except Wiki articles and News. We conjecture this happens because the BERT language model was trained on a large text corpus that comes from these two domains. 
However, importantly, AutoROCC is more robust across domains that are different from these two, since it is an unsupervised approach that is not tuned for any specific domain. 


\begin{table}
    \centering
    \setlength{\tabcolsep}{0.25em}
    \begin{tabular}{@{} l|c|c|c|c @{}}
   \# & Ablations     & ARC & MultiRC & MultiRC   \\
   &  & & EM0 & Justification   \\
   & & & & F1 \\
    \hline
  0 &   Full AutoROCC  & {\bf 56.09} & \bf{25.29} & \bf{56.44}  \\
     \hline
   1 &  -- IDF & 54.11 & 24.65 & 54.19 \\
   2 & -- $C(A)$ & 54.90 & 21.82 & 52.93 \\ 
   3 &  -- $C(Q)$ & 54.66 & 23.61 & 52.09 \\ 
       4 &  -- O  & 55.88 & 24.03 & 55.97 \\
   5 &  R$^\star$  & 53.90  & 23.40 & 44.81 \\
    \end{tabular}
    \caption{Ablation study, removing different components of ROCC. 
    The scores are reported on the ARC test set and MultiRC dev set. R$^\star$ denotes the best approach that relies just on the $R$ score. The hyper parameter $k$ in R$^\star$, was tuned on the development partition of the respective dataset. 
    }
    \label{tab:ablations}
\end{table}

The ARC dataset does not provide justification sentences, so we instead ask how well our question-answering models do on a related inference task, the SciTail entailment dataset \cite{khot2018scitail}.
We trained three QA classifiers on the ARC dataset:
BERT with no justification, BERT with BM25 ($k=4$) justifications,
and BERT with AutoROCC justifications.
We tested these on SciTail, and achieved 64.49\%, 69.70\%, and 73.46\% accuracy, respectively, indicating that AutoROCC's knowledge aggregation is a valid proxy for entailment.




\subsection{Ablation Analysis}

\Cref{tab:ablations} shows an ablation of the different components of ROCC. Row 0 reports the score from the full AutoROCC model.
In row 1, we remove IDF weights from coverage calculations (see \cref{Coverage_IDF}) of both question and answer text. In row 2, 3 and 4, we remove the coverage of answer, coverage of question, and overlap from the ROCC formula (see \cref{ROCC_formula}) respectively. In all the cases, we found small drops in both performance and justification scores across both the datasets, with the removal of either $C(A)$ or $C(Q)$ having the largest impact.


\subsection{Error Analysis}

We analyzed ROCC's justification selection performance on three different types of questions in MultiRC: True/False/Yes/No, Verbatim, and Non-verbatim~\cite{khashabi2018question}. As shown in Table~\ref{tab:error_analysis}, AutoROCC achieves higher recall scores on Verbatim questions, where the answer text is likely to appear within the given justification passage, and worse recall on question types where such overlap does not exist, e.g., Non-verbatim and True/False. This suggests that the $C(A)$ component of ROCC is important for the extraction of meaningful justifications.

\subsection{Alignment ROCC}

To understand the dependence between ROCC and exact lexical match, we compare
the justification selection performance of ROCC when its score components are computed based on lexical match (the approach used throughout the paper up to this point) vs. the semantic alignment match of \citet{yadav2018sanity}. The latter approach relaxes the requirement for lexical match, i.e., two tokens are considered to be matched when the cosine similarity of their embedding vectors is larger than $0.95$.\footnote{This threshold was tuned on the MultiRC development set. We used 100-dimensional GloVe embeddings for this experiment, which performed similarly to larger embedding vectors (300), but allowed for faster experiments.} 
As shown in Table~\ref{tab:lexical_vs_alignment}, the alignment-based ROCC indeed performs better than the ROCC that relies on lexical match. However, the improvements are not large, e.g., the maximum improvement is 1.6\% (when $k=4$), which indicates that ROCC is robust to a certain extent to lexical variation.


\begin{table}
    \centering
    \setlength{\tabcolsep}{0.3em}
    \begin{tabular}{l|c|c|c}
    Question type   & Precision & Recall & F-1 score \\
    \hline
    
    True/False/Yes/No & 54.1 & 68.9 & 60.6 \\
    Verbatim & 49.7 & 71.2 & 58.5 \\
    Non-verbatim & 47.3 & 68.7 & 56.0\\
   
    \end{tabular}
    \caption{Justification selection performance of AutoROCC on different types of questions, in the MultiRC development dataset.}

    \label{tab:error_analysis}
    \vspace{-4mm}
\end{table}

\section{Conclusion}

We introduced ROCC, a simple unsupervised approach for selecting justification sentences for question answering, which balances relevance, overlap of selected sentences, and coverage of the question and answer.
We coupled this method with a state-of-the-art BERT-based supervised question answering system, and achieved a new state-of-the-art on the MultiRC and ARC datasets among approaches that do not use external resources during training.
We showed that ROCC-based QA approaches are more robust across domains, and generalize better to other related tasks like entailment.
In the future, we envision that ROCC scores can be used as distant supervision signal to train supervised justification selection methods.


\begin{table}
    \centering
  
    \begin{tabular}{l|c|c}
    ROCC ($k$ sentences)   & Lexical & Align.  \\
    & ROCC & ROCC \\
    \hline
    
    ROCC ($k = 2$ sentences) & 51.4 & 51.4  \\
    ROCC ($k = 3$ sentences) & 54.7 & 55.5\\
    ROCC ($k = 4$ sentences) & 51.6 & 53.2  \\
    ROCC ($k = 5$ sentences) & 48.1 & 49.2  \\

    \end{tabular}
    \caption{Justification selection performance of the ROCC configuration that uses lexical match (BM25) to retrieve candidate justifications (Lexical ROCC), compared against a ROCC variant that uses the semantic alignment approach of~\citet{yadav2018sanity} to retrieve candidates (Align. ROCC). This experiment used the MultiRC development dataset.} 
  
    \label{tab:lexical_vs_alignment}
    \vspace{-4mm}
\end{table}

\section*{Acknowledgments}
This work was supported by the Defense Advanced Research Projects Agency
(DARPA) under the World Modelers program, grant number
W911NF1810014, and by the National Science Foundation (NSF) under grant IIS-1815948.

Mihai Surdeanu declares a financial interest in lum.ai. This interest has been properly disclosed to the University of Arizona Institutional Review Committee and is managed in accordance with its conflict of interest policies.


\bibliographystyle{acl_natbib}
\bibliography{emnlp-ijcnlp-2019}
\end{document}